\documentclass[10pt,twocolumn,letterpaper]{article}

\usepackage{iccv}
\usepackage{times}
\usepackage{epsfig}
\usepackage{graphicx}
\usepackage{amsmath}
\usepackage{amssymb}
\usepackage{booktabs}

\usepackage[pagebackref=true,breaklinks=true,letterpaper=true,colorlinks,bookmarks=false]{hyperref}

\iccvfinalcopy 

\ificcvfinal\fi

\begin{document}

\title{Hypersim: A Photorealistic Synthetic Dataset for\\Holistic Indoor Scene Understanding}

\author{
Mike Roberts~~~~Jason Ramapuram~~~~Anurag Ranjan~~~~Atulit Kumar\\
Miguel Angel Bautista~~~~Nathan Paczan~~~~Russ Webb~~~~Joshua M. Susskind\\
Apple\\
\url{http://github.com/apple/ml-hypersim}
}

\maketitle

\ificcvfinal\thispagestyle{empty}\fi


\begin{abstract}
\vspace{-0pt}

For many fundamental scene understanding tasks, it is difficult or impossible to obtain per-pixel ground truth labels from real images.
We address this challenge by introducing Hypersim, a photorealistic synthetic dataset for holistic indoor scene understanding.
To create our dataset, we leverage a large repository of synthetic scenes created by professional artists, and we generate 77,400 images of 461 indoor scenes with detailed per-pixel labels and corresponding ground truth geometry.
Our dataset: (1) relies exclusively on publicly available 3D assets; (2) includes complete scene geometry, material information, and lighting information for every scene; (3) includes dense per-pixel semantic instance segmentations and complete camera information for every image; and (4) factors every image into diffuse reflectance, diffuse illumination, and a non-diffuse residual term that captures view-dependent lighting effects.

We analyze our dataset at the level of scenes, objects, and pixels, and we analyze costs in terms of money, computation time, and annotation effort.
Remarkably, we find that it is possible to generate our entire dataset from scratch, for roughly half the cost of training a popular open-source natural language processing model.
We also evaluate sim-to-real transfer performance on two real-world scene understanding tasks -- semantic segmentation and 3D shape prediction -- where we find that pre-training on our dataset significantly improves performance on both tasks, and achieves state-of-the-art performance on the most challenging Pix3D test set.
All of our rendered image data, as well as all the code we used to generate our dataset and perform our experiments, is available online.

\vspace{-0pt}
\end{abstract}

\begin{figure}[t]
\begin{center}
\includegraphics[width=0.48\textwidth]{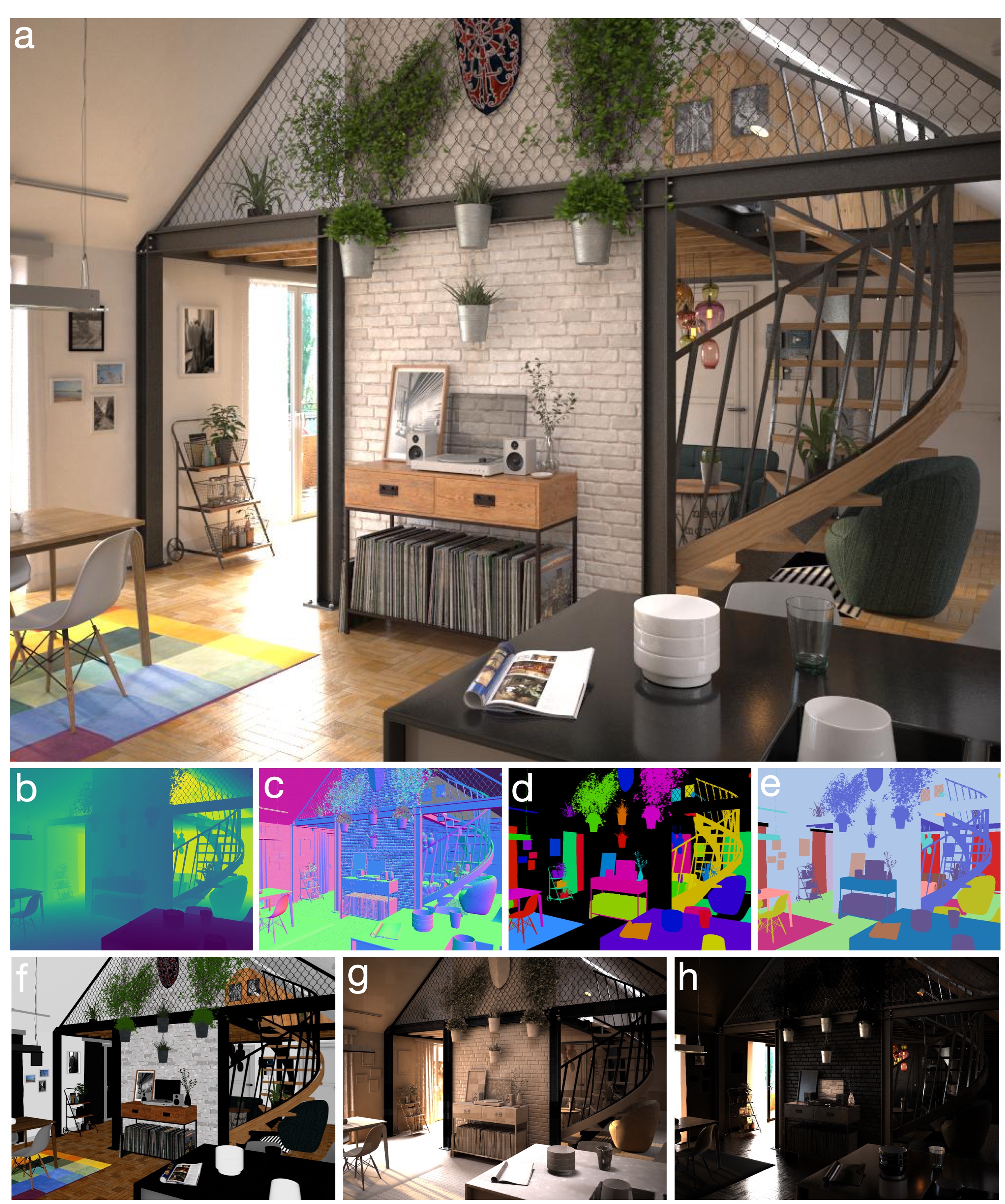}{\vspace{-10pt}}
\end{center}
\caption{
Overview of the Hypersim dataset.
For each color image (a), Hypersim includes the following ground truth layers: depth (b); surface normals (c); instance-level semantic segmentations (d,e); diffuse reflectance (f); diffuse illumination (g); and a non-diffuse residual image that captures view-dependent lighting effects like glossy surfaces and specular highlights (h).
Our diffuse reflectance, diffuse illumination, and non-diffuse residual layers are stored as HDR images, and can be composited together to exactly reconstruct the color image.
}
\vspace{-0pt}
\label{fig:teaser}
\end{figure}

\vspace{-0pt}
\section{Introduction}
\vspace{-0pt}
\label{sec:intro}

\vspace{-5pt}
\begin{table}[t]
\centering
\footnotesize
\begin{tabular}{@{}lllll@{}}
\toprule
Dataset/simulator                                                           & Images     & 3D         & Seg.         & Intrinsic    \\
\midrule
\vspace{2pt}
\emph{Real (3D reconstruction)}                                             &            &            &              &              \\
SceneNN \cite{hua:2016}                                                     &            & \checkmark & S+I          &              \\
Stanford 2D-3D-S \cite{armeni:2017,armeni:2016,xia:2018}                    & \checkmark & \checkmark & S+I          &              \\
Matterport3D \cite{anderson:2018,chang:2017,savva:2017,savva:2019,xia:2018} & \checkmark & \checkmark & S+I          &              \\
ScanNet \cite{dai:2017}                                                     & \checkmark & \checkmark & S+I          &              \\
Gibson \cite{savva:2019,xia:2018}                                           &            & \checkmark &              &              \\
Replica \cite{savva:2019,straub:2019}                                       &            & \checkmark & S+I          &              \\
\midrule
\vspace{2pt}
\emph{Synthetic (artist-created)}                                           &            &            &              &              \\
AI2-THOR \cite{kolve:2017}                                                  &            & \checkmark & S+I          &              \\
ARAP \cite{bonneel:2017}                                                    & \checkmark & \checkmark & I            & D+R          \\
SceneNet-RGBD \cite{mccormac:2017}                                          & \checkmark & \checkmark & S+I          &              \\
PBRS \cite{song:2017,zhang:2017}                                            &            &            & S+I          &              \\
CGIntrinsics \cite{li:2018b,song:2017}                                      & \checkmark &            &              & D            \\
InteriorNet \cite{li:2018a}                                                 & \checkmark &            & S+I          & D            \\
Jiang et al.~\cite{jiang:2018}                                              &            &            & S+I          &              \\
RobotriX \cite{garcia:2018}                                                 & \checkmark & \checkmark & S+I          &              \\
CG-PBR \cite{sengupta:2019,song:2017}                                       &            &            & S            & D+R          \\
DeepFurniture \cite{liu:2019}                                               & \checkmark &            & S            & D            \\
Structured3D \cite{zheng:2019}                                              & \checkmark &            & S+I          & D            \\
TartanAir \cite{shah:2017,wang:2020}                                        & \checkmark &            & I            &              \\
Li et al.~\cite{li:2020a,song:2017}                                         &            &            &              & D+R          \\
3D-FUTURE \cite{fu:2020}                                                    & \checkmark & \checkmark & S+I          &              \\
OpenRooms \cite{li:2020b}                                                   & \checkmark & \checkmark &              & D+R          \\
\textbf{Hypersim (ours)}                                                    & \checkmark & \checkmark & \textbf{S+I} & \textbf{D+R} \\
\bottomrule
\end{tabular}
\normalsize
\vspace{5pt}
\caption{
Comparison to previous datasets and simulators for indoor scene understanding.
We broadly categorize these datasets and simulators as being either \emph{real} (i.e., based on 3D triangle mesh reconstructions from real sensors) or \emph{synthetic} (i.e., artist-created), and we sort chronologically within each category.
We limit our comparisons to synthetic datasets and simulators that aim to be photorealistic. 
The \emph{Images} and \emph{3D} columns indicate whether or not images and 3D assets (e.g., triangle meshes) are publicly available.
The \emph{Seg.}~column indicates what type of segmentation information is available: S indicates semantic; I indicates instance.
The \emph{Intrinsic} column indicates how images are factored into disentangled lighting and shading components: D indicates that each image is factored into diffuse reflectance and diffuse illumination; D+R indicates that each factorization additionally includes a non-diffuse residual term that captures view-dependent lighting effects.
Our dataset is the first to include images, 3D assets, semantic instance segmentations, and a disentangled image representation.
\vspace{-5pt}
}
\label{tbl:comparison}
\end{table}
%
%
%
%
For many fundamental scene understanding tasks, it is difficult or impossible to obtain per-pixel ground truth labels from real images.
In response to this challenge, the computer vision community has developed several photorealistic synthetic datasets and interactive simulation environments that have spurred rapid progress towards the goal of holistic indoor scene understanding \cite{alhashim:2018,anderson:2018,armeni:2017,armeni:2016,bonneel:2017,chang:2017,dai:2017,fu:2020,garcia:2018,handa:2014,hua:2016,jiang:2018,jin:2020,kolve:2017,li:2018a,li:2020a,li:2018b,li:2020b,liu:2019,mccormac:2017,savva:2017,savva:2019,sengupta:2019,shah:2017,song:2017,straub:2019,wang:2019,wang:2020,xia:2018,zhang:2017,zheng:2019}.

However, existing synthetic datasets and simulators have important limitations (see Table \ref{tbl:comparison}).
First, most synthetic datasets are derived from 3D assets that are not publicly available.
These datasets typically include rendered images, but do not include the underlying 3D assets used during rendering (e.g., triangle meshes), and are therefore not suitable for geometric learning problems that require direct 3D supervision (e.g., \cite{gkioxari:2019}).
Second, not all synthetic datasets and simulators include semantic segmentations.
Although it is common for synthetic datasets to include \emph{\emph{some}} kind of segmentation information, these segmentations may not include semantic labels, and may group pixels together at the granularity of low-level object parts, rather than semantically meaningful objects.
Third, most datasets and simulators do not factor images into disentangled lighting and shading components, and are therefore not suitable for inverse rendering problems (e.g., \cite{bell:2014,kovacs:2017}).
No existing synthetic dataset or simulator addresses all of these limitations, including those that target outdoor scene understanding \cite{angus:2018,dosovitskiy:2017,gaidon:2016,hurl:2019,khan:2019,krahenbuhl:2018,richter:2016,richter:2017,ros:2016,sadat:2018,shah:2017,wang:2020,wrenninge:2018}.

In this work, we introduce Hypersim, a photorealistic synthetic dataset for holistic indoor scene understanding that addresses all of the limitations described above (see Figure \ref{fig:teaser}).
To create our dataset, we leverage a large repository of synthetic scenes created by professional artists, and we generate 77,400 images of 461 indoor scenes with detailed per-pixel labels and corresponding ground truth geometry.
Our dataset: (1) relies exclusively on publicly available 3D assets; (2) includes complete scene geometry, material information, and lighting information for every scene; (3) includes dense per-pixel semantic instance segmentations and complete camera information for every image; and (4) factors every image into diffuse reflectance, diffuse illumination, and a non-diffuse residual term that captures view-dependent lighting effects.
Together, these features make our dataset well-suited for geometric learning problems that require direct 3D supervision (e.g., \cite{gkioxari:2019}), multi-task learning problems that require reasoning jointly over multiple input and output modalities (e.g., \cite{standley:2019}), and inverse rendering problems (e.g., \cite{bell:2014,kovacs:2017}).

\begin{figure*}[t]
\begin{center}
\includegraphics[width=0.80\textwidth]{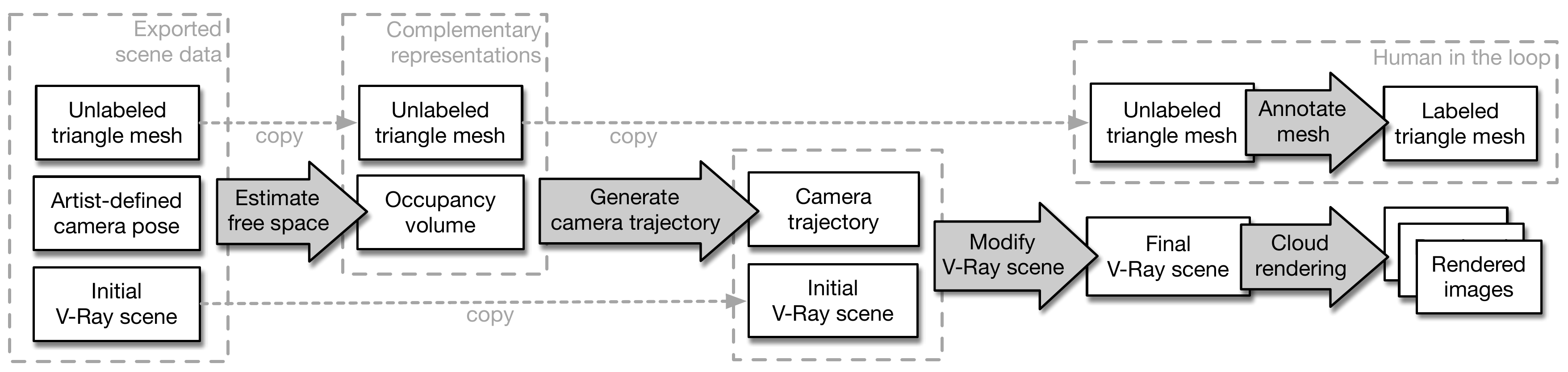}{\vspace{-10pt}}
\end{center}
\caption{
Overview of our computational pipeline.
In this simplified diagram, our pipeline takes as input a triangle mesh, an artist-defined camera pose, and a V-Ray scene description file, and produces as output a collection of images with ground truth labels and corresponding geometry.
The main steps of our pipeline are as follows. We estimate the free space in our scene, use this estimate to generate a collision-free camera trajectory, modify our V-Ray scene to include the trajectory, and invoke our cloud rendering system to render images.
In parallel with the rest of our pipeline, we annotate the scene's triangle mesh using our interactive tool.
In a post-processing step, we propagate mesh annotations to our rendered images (not shown).
This pipeline design enables us to render images before mesh annotation is complete, and also enables us to re-annotate our scenes (e.g., with a different set of labels) without needing to re-render images.
}
\vspace{-5pt}
\label{fig:pipeline}
\end{figure*}

To generate our dataset, we introduce a novel computational pipeline that takes as input a collection of scenes downloaded from an online marketplace, and produces as output a collection of images with ground truth labels and corresponding geometry (see Figure \ref{fig:pipeline}).
Our pipeline has three main steps.
First, we generate camera views of each input scene using a novel view sampling heuristic that does not require the scene to be semantically labeled.
Second, we generate images using a cloud rendering system that we built on top of publicly available cloud computing services.
Third, we obtain semantic segmentations from a human annotator using an interactive mesh annotation tool we built ourselves.

We analyze our dataset at the level of scenes, objects, and pixels, and we analyze costs in terms of money, computation time, and annotation effort.
Remarkably, we find that it is possible to generate our entire dataset from scratch, for roughly half the cost of training a popular open-source natural language processing model.
We also evaluate sim-to-real transfer performance on two scene understanding tasks -- semantic segmentation on NYUv2 \cite{silberman:2012} and 3D shape prediction on Pix3D \cite{sun:2017} -- where we find that pre-training on our dataset significantly improves performance on both tasks, and achieves state-of-the-art performance on the most challenging Pix3D test set.
All of our rendered image data, as well as all the code we used to generate our dataset and perform our experiments, is available online.\footnote{\url{http://github.com/apple/ml-hypersim}}

\vspace{-4pt}
\section{Related Work}
\vspace{-2pt}

\paragraph{Synthetic Data in Computer Vision}

Synthetic data plays a critical role in a wide variety of computer vision applications.
See the excellent recent survey by Nikolenko \cite{nikolenko:2019}.

\vspace{-12pt}
\paragraph{Synthetic Data for Indoor Scene Understanding}

We discuss photorealistic datasets and simulation environments for indoor scene understanding in Section \ref{sec:intro}.
Non-photorealistic datasets and environments \cite{handa:2016b,handa:2016a,song:2017,wu:2018} also play an important role in scene understanding research because they can be rendered very efficiently.
However, these datasets and environments introduce a large domain gap between real and synthetic images that must be handled carefully \cite{nikolenko:2019}.
In contrast, our dataset aims to be as photorealistic as possible, thereby significantly reducing this domain gap.

Several datasets provide 3D CAD models that have been aligned to individual objects in real images \cite{dai:2017,lim:2013,sun:2017,xiang:2016,xiang:2014}. 
In these datasets, the CAD models may not be exactly aligned to each image, and many objects that are visible in an image do not have a corresponding CAD model.
In contrast, our dataset provides segmented 3D models that are exactly aligned to each image, and every pixel of every image is associated with a 3D model.

Shi et al.~\cite{shi:2017} provide photorealistic renderings of individual objects in the ShapeNet dataset \cite{chang:2015}, where each image is factored into diffuse reflectance, diffuse illumination, and a non-diffuse residual term.
Our images are factored in the same way, but we render entire scenes, rather than individual objects.

\vspace{-12pt}
\paragraph{View Sampling Methods for Synthetic Scenes}

Existing view sampling methods for synthetic scenes include: sampling views uniformly at random \cite{wang:2020,zheng:2019}; using data-driven methods to generate realistic camera jitter \cite{li:2018a}; and using semantic segmentation information to match the distribution of semantic classes to an existing dataset \cite{genova:2017}, prefer foreground semantic classes \cite{handa:2016b,handa:2016a,song:2017}, and maintain a minimum number of semantic instances in each view \cite{mccormac:2017,zhang:2017}.
However, existing methods are not directly applicable in our setting.
First, many of our scenes are highly \emph{staged}, i.e., they contain realistic clutter in some parts of the scene, but are unrealistically empty in other parts.
As a result, for our scenes, sampling views uniformly at random produces many uninformative views with no foreground objects.
Second, our pipeline supports rendering images and annotating scenes in parallel, and therefore we do not have access to semantic segmentation information when we are sampling views.
Our view sampling method addresses these challenges by choosing salient views without relying on segmentation information, and is therefore directly applicable in our setting.

\vspace{-12pt}
\paragraph{Interactive Annotation Tools for 3D Scenes}

Our interactive tool is similar in spirit to existing tools for annotating reconstructed triangle meshes \cite{dai:2017,nguyen:2017,straub:2019} and synthetic images from video games \cite{angus:2018,richter:2016}.
All of these tools, including ours, leverage some kind of pre-segmentation strategy to reduce annotation effort.
Tools for annotating reconstructed meshes \cite{dai:2017,nguyen:2017,straub:2019} obtain pre-segmentations by applying an unsupervised grouping method (e.g., \cite{felzenszwalb:2004}) to the input mesh.
As a result, the quality of 2D semantic labels obtained from these tools is limited by the quality of the reconstructed input mesh, as well as the unsupervised grouping.
On the other hand, tools for annotating video game images \cite{angus:2018,richter:2016} obtain clean pre-segmentations by analyzing per-pixel rendering metadata, but do not allow freeform 3D navigation through the scene.
In contrast, our tool leverages clean artist-defined mesh pre-segmentations, leading to 2D semantic labels that are exactly aligned with our rendered images, and our tool allows freeform 3D navigation.

%

\begin{figure*}[t]
\begin{center}
\includegraphics[width=0.95\textwidth]{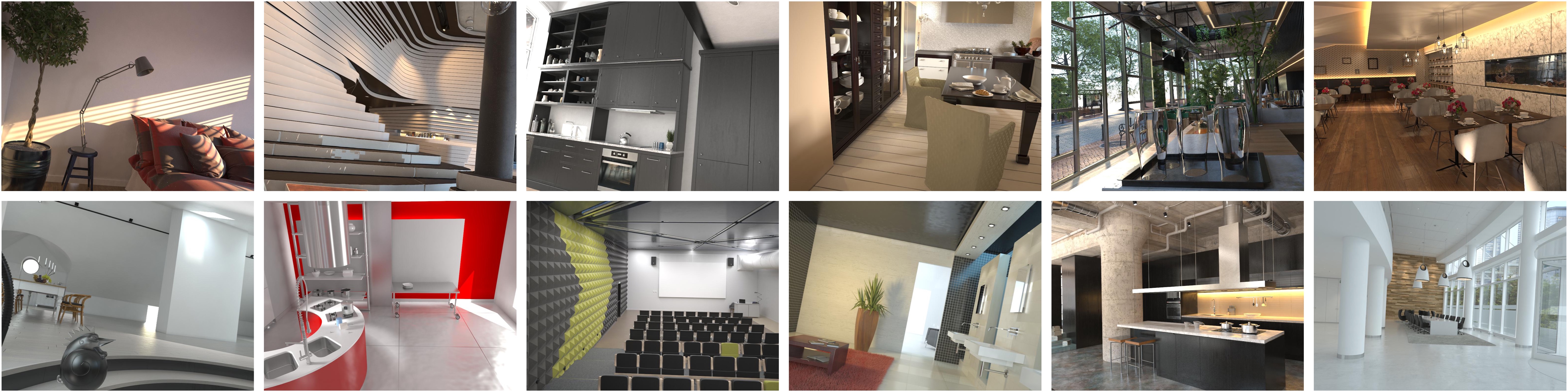}{\vspace{-10pt}}
\end{center}
\caption{
Randomly selected images from our dataset.
From these images, we see that the scenes in our dataset are visually diverse, and our view sampling heuristic generates informative views without requiring our scenes to be semantically labeled.
}
\vspace{-5pt}
\label{fig:random_images}
\end{figure*}

\vspace{-4pt}
\section{Data Acquisition Methodology}
\vspace{-2pt}

To assemble an appropriate collection of scenes for our dataset, we browsed through online marketplaces looking for ready-made indoor scenes that satisfied three main desiderata.
First, we wanted as many scenes as possible.
Second, we wanted scenes that are as photorealistic and visually diverse as possible.
Because our goal in this work is to construct a static dataset, rather than an interactive environment, we are willing to sacrifice rendering speed to achieve larger scale and greater photorealism.
Third, we wanted scenes that are as consistent as possible in terms of their file formats and internal data representations.
This last criterion arises because it is easier to implement our automated computational pipeline if we can parse, modify, and render our scenes in a consistent way.

Motivated by these desiderata, we chose the Evermotion Archinteriors Collection \cite{evermotion:2020} as our starting point\footnote{We purchased the Evermotion Archinteriors Collection from TurboSquid \cite{turbosquid:2020a}.}.
This collection consists of over 500 photorealistic indoor scenes, and has several features that are especially helpful for our purposes.
First, each scene is represented as a standalone asset file that is compatible with V-Ray \cite{vray:2020}.
This representation is helpful because V-Ray has a powerful Python API for programmatically manipulating scenes.
Second, each scene is scaled appropriately in metric units, and has a consistent up direction.
Third, each scene is grouped into object parts, and includes a small set of artist-defined camera poses (i.e., typically between 1 and 5) that frame the scene in an aesthetically pleasing way.
In our pipeline, we use the grouping into object parts as a pre-segmentation that reduces annotation effort, and we use the artist-defined camera poses as a form of weak supervision in several of our processing steps.
Fourth, almost every scene is distributed under a permissive license that allows publicly releasing rendered images, e.g., in academic publications and public benchmarks.

We excluded scenes from our dataset according to the following criteria.
First, we excluded any scenes that depict isolated objects, rather than complete environments.
Second, we excluded any scenes that are not distributed under a royalty-free license.
Third, for each scene, we manually rendered test images, attempted to export mesh data from the standalone asset file, and attempted to generate camera trajectories through the scene using our view sampling heuristic.
We excluded any scenes with noticeable rendering artifacts, any scenes where we were unable to successfully export mesh data, and any scenes where our view sampling heuristic failed to generate views inside the scene's intended viewing region.
After applying these exclusion criteria, we were left with 461 scenes (568 scenes in the Evermotion Archinteriors collection, 107 scenes excluded, 21 of these were because our view sampling heuristic failed).
In our public code release, we provide a complete list of every scene in our dataset.

\begin{figure*}[t]
\begin{center}
\includegraphics[width=0.95\textwidth]{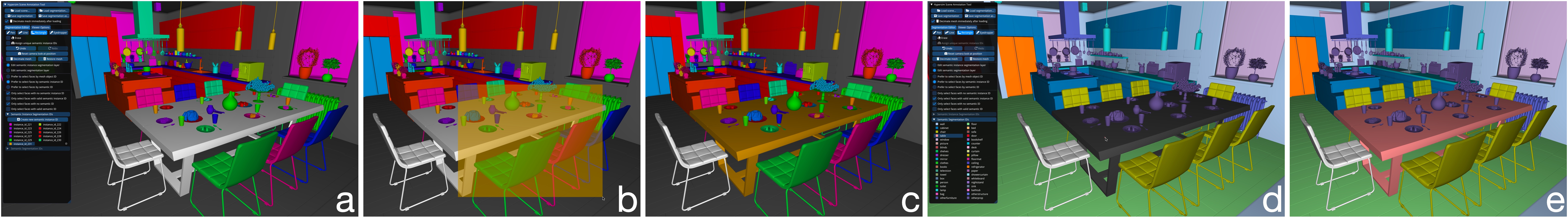}{\vspace{-10pt}}
\end{center}
\caption{
Our interactive mesh annotation tool.
Our tool has a \emph{semantic instance view} (a,b,c) and a \emph{semantic label view} (d,e), as well as a set of \emph{selection filters} that can be used to limit the extent of editing operations based on the current state of the mesh.
To see how these filters can be useful, consider the following scenario.
The table in this scene is composed of multiple object parts, but initially, these object parts have not been grouped into a semantic instance (a).
Our filters enable the user to paint the entire table by drawing a single rectangle, without disturbing the walls, floor, or other objects (b,c).
Once the table has been grouped into an instance, the user can then apply a semantic label with a single button click (d,e).
Parts of the mesh that have not been painted in either view are colored white (e.g., the leftmost chair).
Parts of the mesh that have not been painted in the current view, but have been painted in the other view, are colored dark grey, (e.g., the table in (d)).
Our tool enables the user to accurately annotate an input mesh with very rough painting gestures.
}
\vspace{-5pt}
\label{fig:tool}
\end{figure*}

\begin{figure}[t]
\begin{center}
\includegraphics[width=0.45\textwidth]{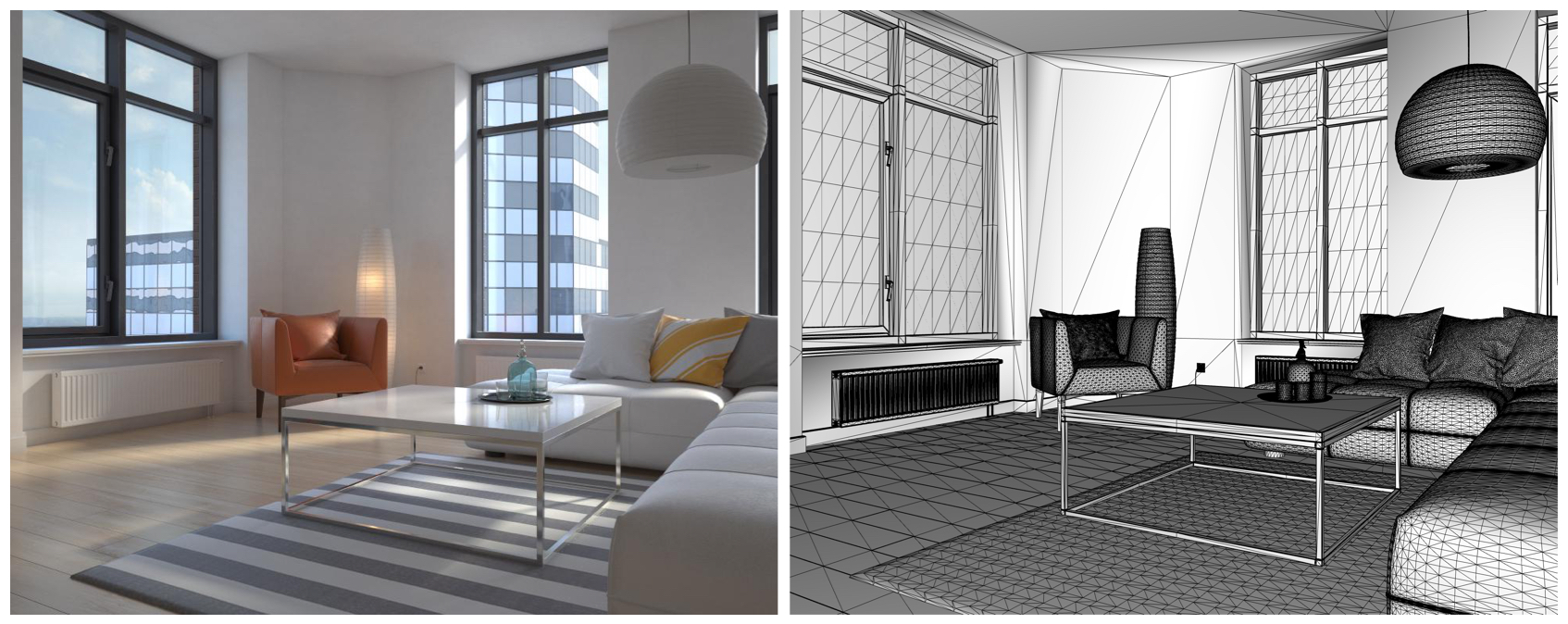}{\vspace{-10pt}}
\end{center}
\caption{
Color and wireframe renderings of a typical scene.
In the wireframe rendering, we observe that the salient objects (e.g., chair, couch, lamp) are more finely tesselated than the non-salient objects (e.g., walls, floor, ceiling).
This observation motivates our view sampling heuristic that takes triangle density into account, and does not require an input scene to be semantically labeled.
}
\vspace{-0pt}
\label{fig:wireframe}
\end{figure}

\begin{figure}[t]
\begin{center}
\includegraphics[width=0.45\textwidth]{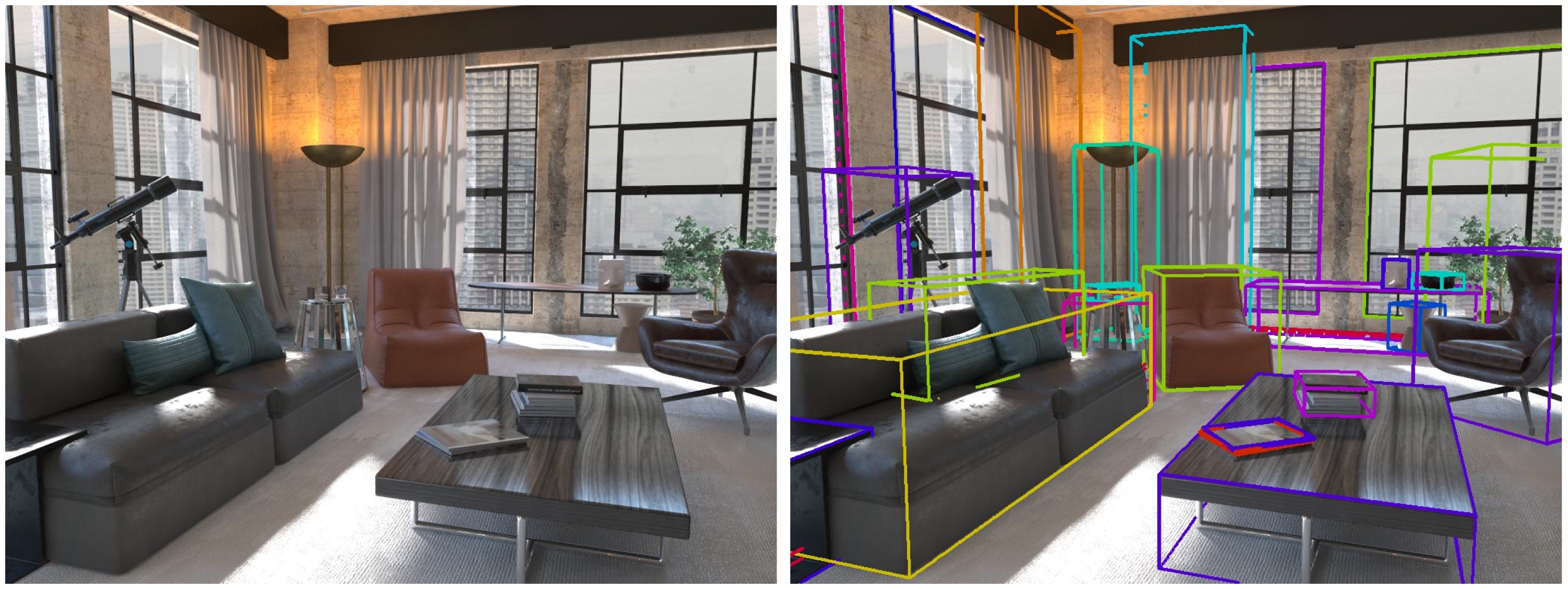}{\vspace{-10pt}}
\end{center}
\caption{
We include a tight 9-DOF bounding box for each semantic instance, so that our dataset can be applied directly to 3D object detection problems (e.g., \cite{song:2015}).
}
\vspace{-5pt}
\label{fig:bounding_box}
\end{figure}

\vspace{-4pt}
\section{Computational Pipeline}
\vspace{-2pt}
\label{sec:pipeline}

After acquiring our scenes, we apply our computational pipeline to generate images with ground truth labels and corresponding geometry (see Figure \ref{fig:pipeline}).
In this section, we describe our computational pipeline, and we assume for simplicity that we are processing a single scene.
To generate our full dataset, we apply the same pipeline to all of our scenes.
For any pipeline steps that require manual data filtering, we provide a complete record of our filtering decisions in our public code release, so our dataset can be reproduced exactly.
In the supplementary material, we describe  our procedure for estimating free space, our procedure for modifying V-Ray scenes, and our cloud rendering system.
\vspace{-12pt}
\paragraph{Pre-processing}

We assume that we are given as input a standalone asset file that describes our scene.
We begin our pipeline by programmatically exporting a triangle mesh, all artist-defined camera poses, and a V-Ray scene description file from the original asset file.
We manually remove from our dataset any artist-defined camera poses that are outside the intended viewing region for the scene.

\textit{Output.}
In total, we exported 784 artist-defined camera poses (809 exported initially, 25 removed manually).

%

\vspace{-12pt}
\paragraph{Generating Camera Trajectories}

We generate camera trajectories using a simple view sampling heuristic that does not require the input scene to be semantically labeled, works well for our scenes, and has not previously appeared in the literature.
When designing our heuristic, we make the observation that salient objects (e.g., chairs, couches, lamps) tend to be more finely tesselated than non-salient objects (e.g., walls, floors, ceilings) (see Figure \ref{fig:wireframe}).
Motivated by this observation, we define a \emph{view saliency model} that takes triangle density into account, and we sample views based on this model.
In our model, we also include a term that penalizes views for observing empty pixels, i.e., pixels that do not contain any scene geometry.

Stating our approach formally, we define the view saliency $v(\mathbf{c})$ of the camera pose $\mathbf{c}$ as follows,
\begin{equation}
v(\mathbf{c}) = t^{\alpha}p^\beta
\end{equation}
where
$t$ is the number of unique triangles observed by $\mathbf{c}$;
$p$ is the fraction of non-empty pixels observed by $\mathbf{c}$;
and $\alpha,\beta > 0$ are parameters that control the sensitivity of our model to triangle counts and empty pixels, respectively.

Using our view saliency model, we generate camera trajectories by constructing random walks through free space that begin at each artist-defined camera pose, and are biased towards upright salient views.
In Figure \ref{fig:random_images}, we show randomly selected images from our dataset that were generated according to this sampling procedure.
We include our exact random walk formulation in the supplementary material.
Our trajectories can occasionally drift outside the scene's intended viewing region, e.g., through an open window.
To address this issue, we manually remove any such trajectories from our dataset.

\textit{Output.}
In total, we generated 774 camera trajectories using our random walk sampling approach (784 generated initially, 10 removed manually).
In order to achieve reasonable visual coverage of our scenes while maintaining tolerable rendering costs, we defined each trajectory to consist of 100 camera poses, ultimately leading to a total of 77,400 distinct views that must be rendered.

\vspace{-12pt}
\paragraph{Interactive Mesh Annotation}

\begin{figure}[t]
\begin{center}
\includegraphics[width=0.48\textwidth]{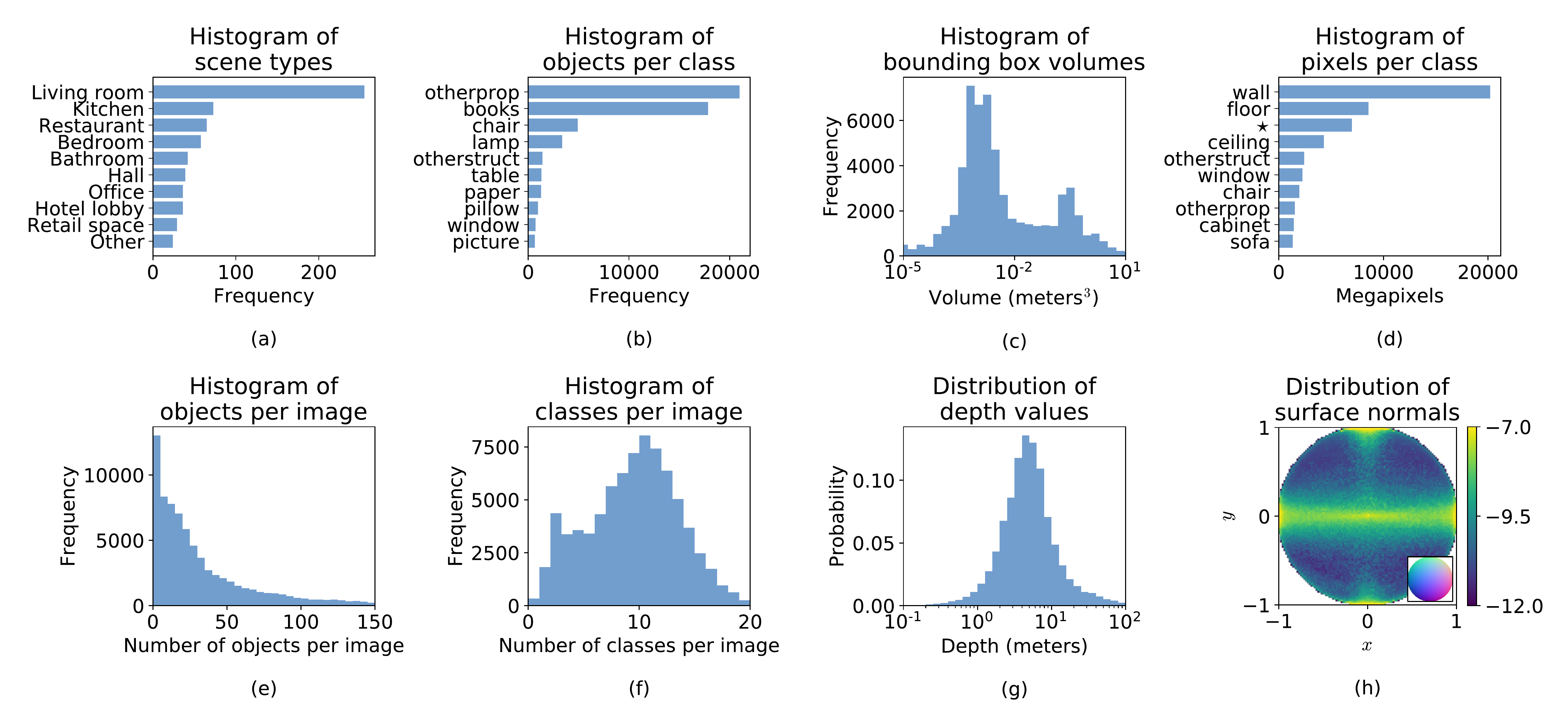}{\vspace{-15pt}}
\end{center}
\caption{
Dataset statistics at the granularity of scenes (a), objects (b,c), and pixels (d,e,f,g,h).
We truncate the histograms in (a,b,d) to show the 10 most common scene types and object classes.
In (a), we assign a scene type to each camera trajectory, and we count the number of trajectories belonging to each scene type.
In (b), we count the number of unique objects in each scene that are visible in at least one image.
In (d), $\star$ indicates pixels with no class label.
In (g), we compute the Euclidean distance from the surface at each pixel to the optical center of the camera.
In (h), we show the distribution of camera-space normals as a function of the normal's $x$ and $y$ coordinates (i.e., how much the normal is pointing to the right and up in camera space), where color indicates log-probability, and the small inset indicates our mapping from normal values to RGB values (e.g., in Figure \ref{fig:teaser}c).
The horizontal axes in (c) and (g) are log-scaled.
}
\vspace{-7pt}
\label{fig:stats_scenes_objects_images}
\end{figure}

\begin{figure}[t]
\begin{center}
\includegraphics[width=0.45\textwidth]{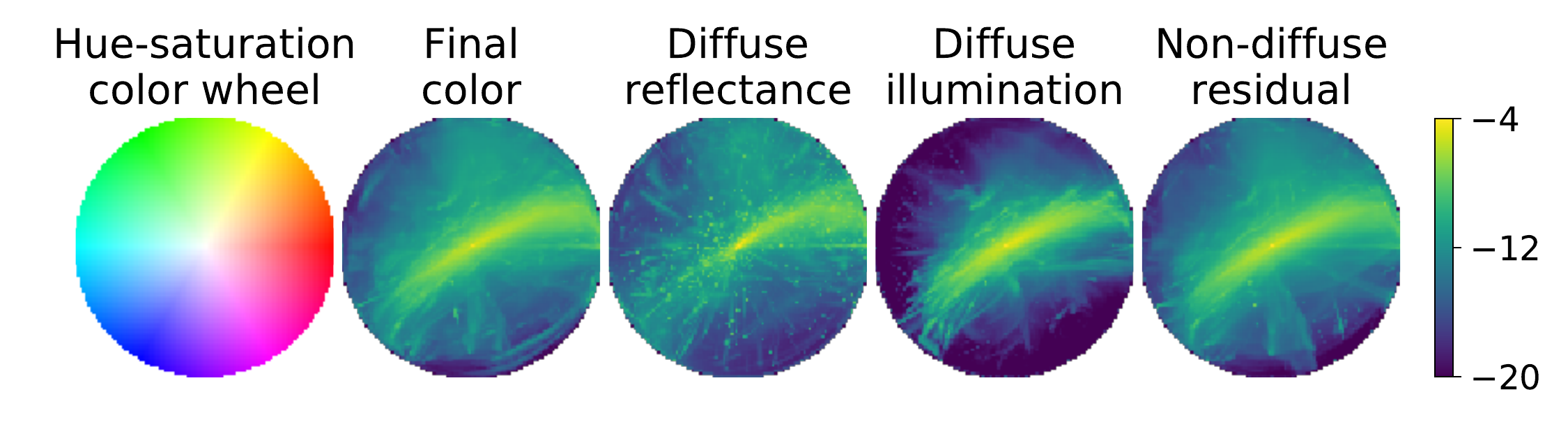}{\vspace{-10pt}}
\end{center}
\caption{
Distributions of hue-saturation values in our distentangled image representation.
We normalize each RGB value independently, convert it to HSV space, and show the resulting hue-saturation distributions.
Color indicates log-probability.
}
\vspace{-7pt}
\label{fig:stats_color_hue_saturation}
\end{figure}

\begin{figure}[t]
\begin{center}
\includegraphics[width=0.45\textwidth]{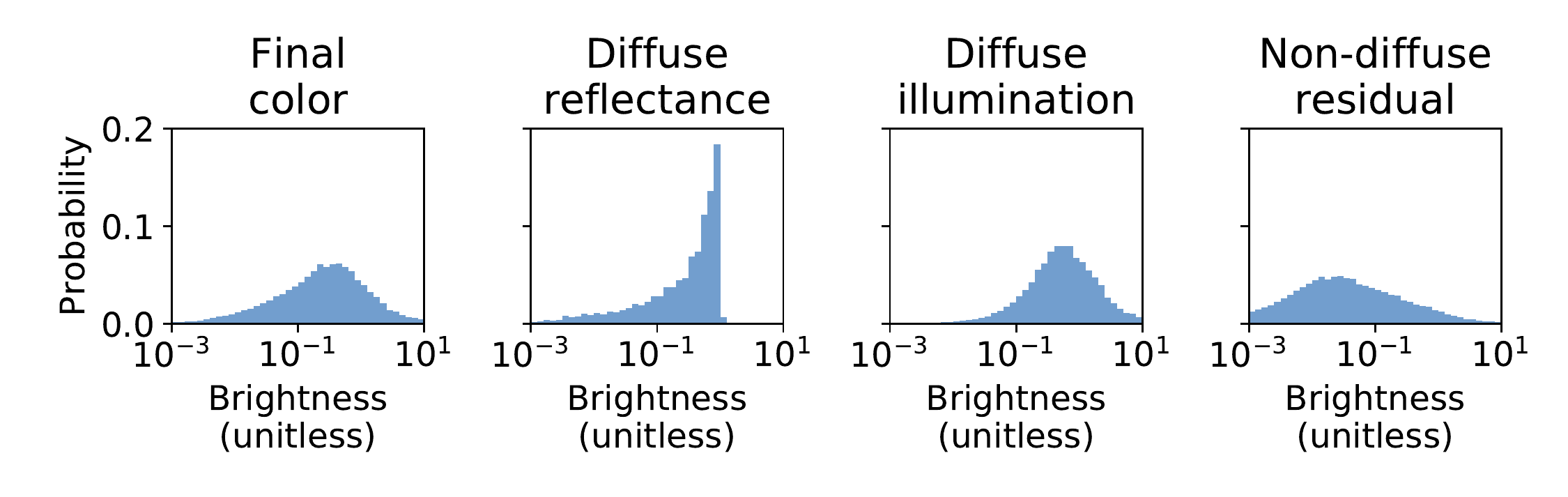}{\vspace{-10pt}}
\end{center}
\caption{
Distributions of brightness values in our distentangled image representation.
We store our image data in an un-normalized HDR format, so we observe brightness values outside the range [0,1].
The horizontal axes in these plots are log-scaled.
}
\vspace{-7pt}
\label{fig:stats_color_brightness}
\end{figure}

\begin{figure}[t]
\begin{center}
\includegraphics[width=0.38\textwidth]{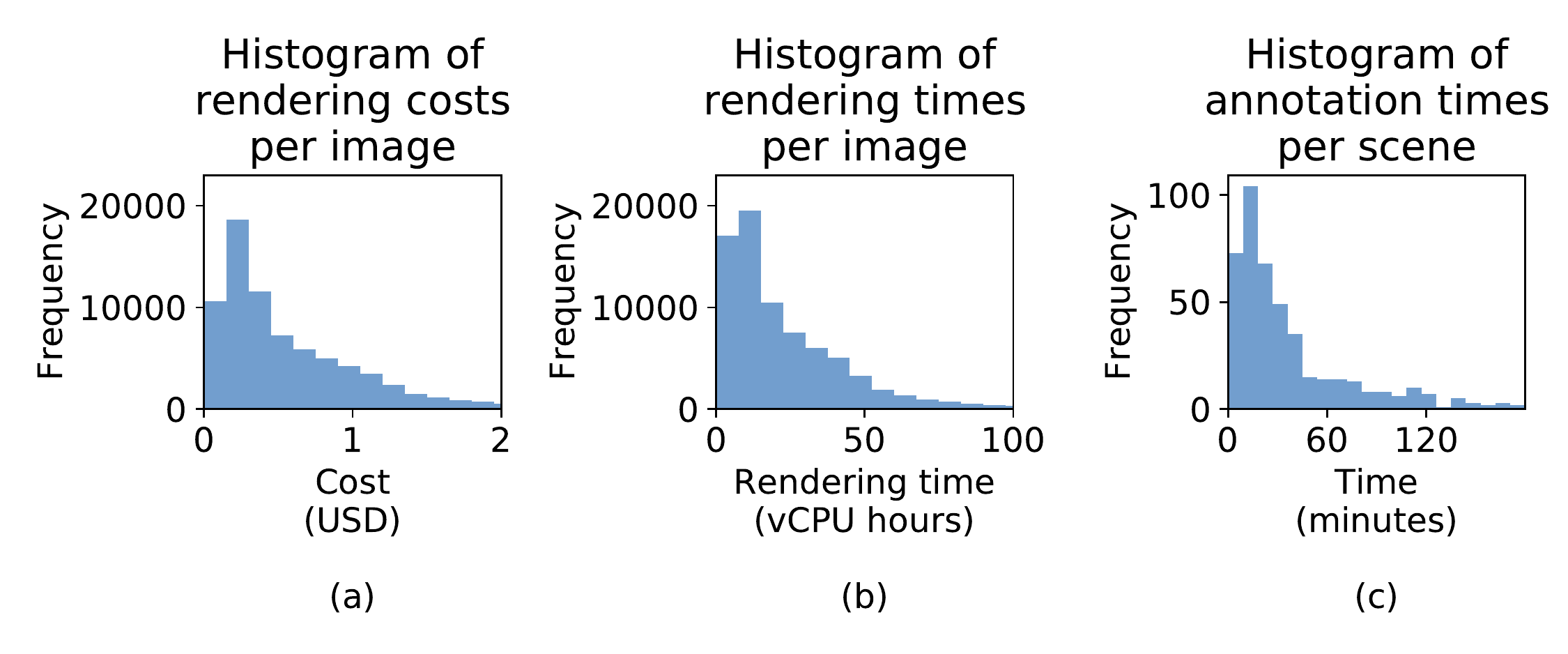}{\vspace{-10pt}}
\end{center}
\caption{
Costs of generating our dataset in terms of money (a), computation time (b), and annotation effort (c).
}
\vspace{-7pt}
\label{fig:stats_costs}
\end{figure}

In parallel with the rest of our pipeline, we obtain semantic segmentations using an interactive mesh annotation tool we built ourselves (see Figure \ref{fig:tool}).
In addition to providing a mesh painting interface that operates at the granularity of object parts \cite{angus:2018,dai:2017,nguyen:2017,richter:2016,straub:2019}, our tool has two unique features that are especially helpful for annotating our scenes.

First, our tool provides a set of \emph{selection filters} that can be used to limit the extent of editing operations based on the current state of the mesh.
These filters enable the user to accurately annotate the mesh using very rough painting gestures (see Figure \ref{fig:tool}).


Second, many of our meshes are large (i.e., millions of triangles) and incoherently laid out in memory.
These meshes cannot be rendered at interactive rates, even on modern GPUs, without the use of advanced acceleration structures \cite{akeninemoller:2018,isenburg:2005,luebke:2002}.
In our tool, we address this issue by including an option to \emph{decimate} the input mesh, which simply removes triangles at random until the mesh fits within a user-specified triangle budget.
We bias our decimation procedure to remove very small triangles, and optionally triangles that are very far away from the camera, since removing these triangles tends to have a negligible impact on the user.
Our tool guarantees that annotations on the decimated mesh are implicitly propagated back to the full-resolution mesh.

\textit{Output.}
Using our tool, we annotated our entire dataset of 461 scenes with instance-level NYU40 labels \cite{gupta:2013,silberman:2012}.

\vspace{-12pt}
\paragraph{Post-processing}

In a final post-processing step, we propagate our mesh annotations to our images using per-pixel metadata that we generate during rendering. See the supplementary material for details.
We also generate a tight 9-DOF bounding box for each semantic instance \cite{barequet:2001,malandain:2002}, so that our dataset can be applied directly to 3D object detection problems (e.g., \cite{song:2015}) (see Figure \ref{fig:bounding_box}).

\vspace{-4pt}
\section{Analysis}
\vspace{-2pt}

In this section, we analyze our dataset at the level of scenes, objects, and pixels, and we analyze costs in terms of money, computation time, and annotation effort.
We summarize the contents of our dataset in Figures \ref{fig:stats_scenes_objects_images}, \ref{fig:stats_color_hue_saturation}, \ref{fig:stats_color_brightness}, and we summarize costs in Figure \ref{fig:stats_costs}.

\vspace{-12pt}
\paragraph{Scenes}

Our dataset consists mostly of residential scenes (e.g., living rooms, kitchens, bedrooms, bathrooms), but commercial scenes (e.g., offices, restaurants) are also common (Fig.~\ref{fig:stats_scenes_objects_images}a).
Our scenes are highly cluttered, containing 127.3 objects per scene on average.
This level of clutter and annotation detail compares favorably to existing indoor scene datasets (e.g., ScanNet \cite{dai:2017}: 14 objects per scene; Replica \cite{straub:2019}: 84 objects per scene).
Our scenes range in size from $<$0.1 up to 11 million triangles.

\vspace{-12pt}
\paragraph{Objects}

At the object level, the most common semantic classes in our dataset are \{\emph{otherprop}, \emph{books}, \emph{chair}\} (Fig.~\ref{fig:stats_scenes_objects_images}b).
The prevalence of these classes is roughly consistent with existing indoor scene datasets \cite{dai:2017,song:2015,straub:2019}.
The distribution of bounding box volumes for our objects is bi-modal, where the two modes correspond to the bounding box volumes of a coffee mug and an office chair (Fig.~\ref{fig:stats_scenes_objects_images}c).

\vspace{-12pt}
\paragraph{Segmentation Images}

At the pixel level, the most common semantic classes in our dataset are \{\emph{wall}, \emph{floor}, \emph{ceiling}\} (Fig.~\ref{fig:stats_scenes_objects_images}d), which tend to dominate even in cluttered images (Fig.~\ref{fig:teaser}e).
88.3\% of pixels in our dataset have a semantic class label, and 52.0\% of pixels have a semantic instance ID.
This annotation density is lower than GTA5 \cite{richter:2016} (98.3\%), but higher than ScanNet \cite{dai:2017} (76\%).

The images in our dataset contain 8.9 classes on average, 49.9 objects on average, and 51.5\% of our images contain 21 or more objects (Fig.~\ref{fig:stats_scenes_objects_images}e,f).
This level of clutter and annotation detail compares favorably to existing semantic segmentation datasets (e.g., NYUv2 \cite{silberman:2012}: 23.5 objects per image; COCO \cite{lin:2014}: 7.2 objects per image; ADE20K \cite{zhou:2019}: 19.6 objects per image), and offers quantitative evidence that our view sampling heuristic is successfully generating informative views of our scenes.

\vspace{-12pt}
\paragraph{Depth Images}

Our depth values are log-normally distributed with an average depth of 5.4 meters (Fig.~\ref{fig:stats_scenes_objects_images}g).
This distribution is roughly consistent with existing indoor depth datasets \cite{silberman:2012,vasiljevic:2019,xiao:2013}, but is noticeably different from outdoor datasets \cite{cordts:2016,geiger:2013}.

\vspace{-12pt}
\paragraph{Surface Normal Images}

Our distribution of surface normals is biased towards planar surfaces, e.g., walls, floors, and ceilings viewed from an upright camera (Fig.~\ref{fig:stats_scenes_objects_images}g).
This distribution disagrees with the \emph{surface isotropy prior} used in the popular SIRFS model \cite{barron:2015} for decomposing images into disentangled shape, lighting, and shading components.
The authors of SIRFS note that surface isotropy is an appropriate prior for object-centric tasks, but is less appropriate for scene-centric tasks, and our data supports this assertion. 

\vspace{-12pt}
\paragraph{Disentangled Lighting and Shading Images}

We observe that some diffuse reflectance values are more likely than others (e.g., desaturated oranges are more likely than saturated greens), and that our distribution of diffuse reflectance is \emph{speckled}, indicating a sparse palette of especially likely values (Fig.~\ref{fig:stats_color_hue_saturation}).
Our distribution of diffuse illumination is especially biased towards natural lighting conditions (Fig.~\ref{fig:stats_color_hue_saturation}).
These distributions roughly agree with common priors on reflectance and illumination found in the inverse rendering literature \cite{barron:2015,bonneel:2017}.

The distributions of hue-saturation values in our non-diffuse residual images and final color images are similar (Fig.~\ref{fig:stats_color_hue_saturation}).
(If our scenes consisted of half perfectly diffuse and half perfectly specular surfaces, we would expect these two distributions to be identical.)
However, our residual images are much more sparse, i.e., close to zero brightness most of the time (Fig.~\ref{fig:teaser}h, Fig.~\ref{fig:stats_color_brightness}).
Nonetheless, we observe that our residual images contribute a non-trivial amount of energy to our final images, and this observation validates our decision to represent non-diffuse illumination explicitly in our dataset.

\begin{table}[t]
\centering
\footnotesize
\begin{tabular}{@{}llcc@{}}
\toprule
\multicolumn{2}{c}{Training procedure} \\
\cmidrule(lr){1-2}
Pre-train       & Fine-tune    & mIoU          & mIoU       \\
                &              & (13-class)    & (40-class) \\
\midrule
None            & NYUv2 (100\%) & 45.2          & 31.4          \\
\cmidrule{1-4}
                & NYUv2 (25\%)  & 46.4          & 29.0          \\
Hypersim (ours) & NYUv2 (50\%)  & 49.1          & 32.7          \\
                & NYUv2 (100\%) & \textbf{51.6} & \textbf{36.4} \\
\bottomrule
\end{tabular}
\normalsize
\vspace{5pt}
\caption{
Sim-to-real performance of our dataset on NYUv2 \cite{silberman:2012} semantic segmentation.
Higher is better.
In parentheses, we show the amount of NYUv2 training data used during training.
For 13-class segmentation, pre-training on our dataset and fine-tuning on 25\% of the NYUv2 training set outperforms training on the full NYUv2 training set.
For 40-class segmentation, pre-training on our dataset and fine-tuning on 50\% of the NYUv2 training set outperforms training on the full NYUv2 training set.
\vspace{-5pt}
}
\label{tbl:segmentation}
\end{table}

\begin{table}[t]
\centering
\footnotesize
\begin{tabular}{@{}llccc@{}}
\toprule
\multicolumn{2}{c}{Training procedure} \\
\cmidrule(lr){1-2}
Pre-train       & Fine-tune & AP$^{\text{mesh}}$ & AP$^{\text{mask}}$ & AP$^{\text{box}}$ \\
\midrule
None            & Pix3D     & 28.8               & 63.9               & 72.2              \\
Hypersim (ours) & Pix3D     & \textbf{29.6}      & \textbf{64.6}      & \textbf{72.7}     \\
\bottomrule
\end{tabular}
\normalsize
\vspace{5pt}
\caption{
Sim-to-real performance of our dataset on Pix3D \cite{sun:2017} 3D shape prediction.
Following \cite{gkioxari:2019}, we report AP$^{\text{mesh}}$, AP$^{\text{mask}}$, and AP$^{\text{box}}$ on the $\mathcal{S}_2$ test set. Higher is better.
On the top row, we show the previous state-of-the-art, achieved by training Mesh R-CNN \cite{gkioxari:2019} on Pix3D. On the bottom row, we show our result, achieved by pre-training Mesh R-CNN on our dataset and fine-tuning on Pix3D. Pre-training on our dataset achieves state-of-the-art performance.
\vspace{-5pt}
}
\label{tbl:completion}
\end{table}

\vspace{-12pt}
\paragraph{Rendering Costs}

In total, it cost \$57K to generate our dataset (\$6K to purchase 461 scenes, \$51K to render 77,400 images), and took 231 vCPU years (2.4 years of wall-clock time on a large compute node).
Although generating our entire dataset is undoubtedly expensive, it is considerably less expensive than some other popular learning tasks.
For example, generating our dataset is 0.56$\times$ as expensive as training the open-source Megatron-LM natural language processing model \cite{shoeybi:2019}, which would cost \$103K to train from scratch using publicly available cloud computing services.

Per-image rendering costs in dollars are linearly related to compute times, and we observe that both are log-normally distributed (Fig.~\ref{fig:stats_costs}a,b).
On average, rendering a single image in our dataset at 1024$\times$768 resolution costs \$0.67 and takes 26 vCPU hours (20 minutes of wall-clock time on a large compute node).
We include the cost of rendering each image in our public code release, so the marginal value and marginal cost of each image can be analyzed jointly in downstream applications \cite{settles:2012}.

\vspace{-12pt}
\paragraph{Annotation Costs}

In total, annotating our scenes took 369 hours, leading to an average annotation speed of 39.8K pixels per second.
Our annotation speed is two orders of magnitude faster than the fine-grained annotations in Cityscapes \cite{cordts:2016} (0.3K pixels per second), but an order of magnitude slower than GTA5 \cite{richter:2016} (279.5K pixels per second).
However, our annotations provide complementary information that isn't available in either of these datasets.
For example, our mesh annotations provide segmentation information for occluded parts of the scene that aren't visible in any image, which could be used for \emph{\emph{amodal}} segmentation problems (e.g., \cite{li:2016}).
Per-scene annotation times are log-normally distributed, with a single scene taking 48 minutes to annotate on average (Fig.~\ref{fig:stats_costs}c).
All of our manual filtering decisions took less than 8 hours.


\vspace{-4pt}
\section{Experiments}
\vspace{-2pt}

In this section, we evaluate the sim-to-real transfer performance of our dataset on semantic segmentation and 3D shape prediction.
For both tasks, our methodology is to pre-train on our dataset, fine-tune on an appropriate real-world dataset, and evaluate performance on the real-world dataset.
We summarize our results in Tables \ref{tbl:segmentation} and \ref{tbl:completion}, and we include additional details in the supplementary material.

\vspace{-12pt}
\paragraph{Data Splits}

In our public code release, we provide standard \{training, validation, test\} splits consisting of \{59,543, ~7,386, ~7,690\} images each.
We split our dataset by scene, i.e., every image of a given scene belongs to the same split.
We use these splits in all our experiments.

\vspace{-12pt}
\paragraph{Semantic Segmentation}

We evaluate semantic segmentation performance on the NYUv2 dataset \cite{silberman:2012}, which consists of 795 training images and 654 test images.
We evaluate performance on the 13-class \cite{couprie:2013} and 40-class \cite{gupta:2013} segmentation tasks, and we report mean intersection-over-union (mIoU) as our evaluation metric.
In all our experiments, we use RGB images for training and testing.

We use a standard U-Net \cite{ronneberger:2015} as our model architecture, with a ResNet-34 \cite{he:2016} encoder initialized  using ImageNet \cite{russakovsky:2015} weights.
During training, we always apply random \{cropping, flipping, resizing\}, and we apply color jittering with probability 0.5.
We use an identical training recipe during pre-training and fine-tuning, and we perform our evaluation at 512$\times$512 resolution.

We find that pre-training on our dataset significantly improves semantic segmentation performance on NYUv2 (see Table \ref{tbl:segmentation}).
Our semantic segmentation results (+6.2 mIoU 13-class; +5.0 mIoU 40-class) are better than the reported results for PBRS \cite{song:2017,zhang:2017} (+1.6 mIoU 40-class), but worse than the reported results for SceneNet-RGBD \cite{mccormac:2017} (+8.1 mIoU 13-class), under similar experimental conditions.
However, our dataset (77K images, 0.5K scenes) is an order of magnitude smaller than PBRS (568K images, 45K scenes), and two orders of magnitude smaller than SceneNet-RGBD (5,000K images, 16K scenes).
Despite its smaller size, we attribute the competitive performance of our dataset to the increased photorealism of our images and scenes.
This finding suggests that there are multiple viable strategies for achieving good sim-to-real performance on a fixed rendering budget -- a small, more photorealistic dataset can be competitive with a large, less photorealistic dataset.
 Determining the optimal \emph{portfolio} of more and less photorealistic images for a given downstream task, subject to a fixed rendering budget,  is an exciting direction for future work \cite{settles:2012}. 

\vspace{-12pt}
\paragraph{3D Shape Prediction}

We evaluate 3D shape prediction performance on the Pix3D dataset \cite{sun:2017}, which consists of 10,069 images and 395 unique triangle meshes.
We use the data splits defined in \cite{gkioxari:2019} for training, validation, and testing, and we perform our final evaluation on the $\mathcal{S}_2$ test set, which is the most challenging Pix3D test set. 
Following \cite{gkioxari:2019}, we report AP$^{\text{mesh}}$, AP$^{\text{mask}}$, and AP$^{\text{box}}$.

We use Mesh R-CNN \cite{gkioxari:2019} as our model architecture.
We follow the authors' training recipe exactly, except we tune the learning rate used during pre-training.
When selecting our learning rate, we follow the authors' guidelines \cite{meshrcnn:2020} for choosing hyperparameters.

\vspace{-4pt}
\section{Conclusions}
\vspace{-2pt}

We leveraged a large repository of synthetic scenes to create a new dataset for holistic indoor scene understanding.
We introduced a novel computational pipeline to generate salient views of our scenes and render photorealistic images in the cloud, as well as a new interactive tool to efficiently annotate our scenes.
We used our pipeline and annotation tool to create the first computer vision dataset to combine images, 3D assets, semantic instance segmentations, and a disentangled image representation.
We analyzed the costs of generating our dataset, and we found that it is possible to generate our entire dataset from scratch, for roughly half the cost of training a popular open-source natural language processing model.
Finally, we demonstrated that pre-training on our dataset improves performance on two real-world scene understanding tasks, and achieves state-of-the-art performance on the most challenging Pix3D\ test set.

We believe our dataset can enable progress on a wide range of computer vision problems where obtaining real-world ground truth is difficult or impossible.
In particular, our dataset is well-suited for geometric learning problems that require 3D supervision, multi-task learning problems, and inverse rendering problems.
Our labeled scenes could be used to train automatic mesh segmentation systems, as well as generative modeling systems that synthesize realistic scene variations on-demand.
Moving beyond our specific dataset, we see many potential applications for photorealistic synthetic data in computer vision, and we believe there are abundant opportunities to co-design rendering algorithms and learning algorithms to amortize rendering costs more effectively.

\vspace{-4pt}
\section*{Acknowledgements}
\vspace{-2pt}

We thank the professional artists at Evermotion for making their Archinteriors Collection available for purchase;
Danny Nahmias for helping us to acquire data;
Max Horton for helping us to prototype our annotation tool;
Momchil Lukanov and Vlado Koylazov at Chaos Group for their excellent support with V-Ray;
David Antler, Hanlin Goh, and Brady Quist for proofreading the paper;
Ali Farhadi, Zhile Ren, Fred Schaffalitzky, and Qi Shan for the helpful discussions;
and Jia Zheng for catching and correcting an error in Table \ref{tbl:comparison}.

\appendix

{
\small
\bibliographystyle{ieee_fullname}
\bibliography{00_preamble}

\begin{thebibliography}{10}\itemsep=-1pt

\bibitem{meshrcnn:2020}
\url{http://github.com/facebookresearch/meshrcnn/issues/31}.

\bibitem{vray:2020}
{Chaos Group V-Ray}.
\newblock \\\url{http://www.chaosgroup.com}.

\bibitem{evermotion:2020}
{Evermotion Archinteriors Collection}.
\newblock \\\url{http://www.evermotion.org}.

\bibitem{turbosquid:2020a}
{TurboSquid}.
\newblock \\\url{http://www.turbosquid.com}.

\bibitem{akeninemoller:2018}
Tomas Akenine-M{\"o}ller, Eric Haines, Naty Hoffman, Angelo Pesce, Micha{\l}
  Iwanicki, , and S{\`e}bastien Hillaire.
\newblock {\em Real-Time Rendering, Fourth Edition}.
\newblock CRC Press, 2018.

\bibitem{alhashim:2018}
Ibraheem Alhashim and Peter Wonka.
\newblock High quality monocular depth estimation via transfer learning.
\newblock arXiv 2018.

\bibitem{anderson:2018}
Peter Anderson, Qi Wu, Damien Teney, Jake Bruce, Mark Johnson, Niko
  S{\"u}nderhauf, Ian Reid, Stephen Gould, and Anton van~den Hengel.
\newblock Vision-and-language navigation: {I}nterpreting visually-grounded
  navigation instructions in real environments.
\newblock In {\em CVPR 2018}.

\bibitem{angus:2018}
Matt Angus, Mohamed ElBalkini, Samin Khan, Ali Harakeh, Oles Andrienko, Cody
  Reading, Steven Waslander, and Krzysztof Czarnecki.
\newblock Unlimited road-scene synthetic annotation ({URSA}) dataset.
\newblock arXiv 2018.

\bibitem{armeni:2017}
Iro Armeni, Alexander Sax, Amir~R. Zamir, and Silvio Savarese.
\newblock Joint {2D}-{3D}-semantic data for indoor scene understanding.
\newblock arXiv 2017.

\bibitem{armeni:2016}
Iro Armeni, Ozan Sener, Amir~R. Zamir, Helen Jiang, Ioannis Brilakis, Martin
  Fischer, and Silvio Savarese.
\newblock {3D} semantic parsing of large-scale indoor spaces.
\newblock In {\em CVPR 2016}.

\bibitem{barequet:2001}
Gill Barequet and Sariel Har-Peled.
\newblock Efficiently approximating the minimum-volume bounding box of a point
  set in three dimensions.
\newblock In {\em ACM-SIAM Symposium on Discrete Algorithms 2001}.

\bibitem{barron:2015}
Jonathan~T. Barron and Jitendra Malik.
\newblock Shape, illumination, and reflectance from shading.
\newblock {\em PAMI}, 2015.

\bibitem{bell:2014}
Sean Bell, Kavita Bala, and Noah Snavely.
\newblock Intrinsic images in the wild.
\newblock In {\em SIGGRAPH 2014}.

\bibitem{bonneel:2017}
Nicolas Bonneel, Balazs Kovacs, Sylvain Paris, and Kavita Bala.
\newblock Intrinsic decompositions for image editing.
\newblock {\em Computer Graphics Forum}, 36(2), 2017.

\bibitem{chang:2017}
Angel Chang, Angela Dai, Thomas Funkhouser, Maciej Halber, Matthias Niessner,
  Manolis Savva, Shuran Song, Andy Zeng, and Yinda Zhang.
\newblock {Matterport3D}: {L}earning from {RGB-D} data in indoor environments.
\newblock In {\em 3DV 2017}.

\bibitem{chang:2015}
Angel~X. Chang, Thomas Funkhouser, Leonidas Guibas, Pat Hanrahan, Qixing Huang,
  Zimo Li, Silvio Savarese, Manolis Savva, Shuran Song, Hao Su, Jianxiong Xiao,
  Li Yi, and Fisher Yu.
\newblock {ShapeNet}: {A}n information-rich {3D} model repository.
\newblock arXiv 2015.

\bibitem{cordts:2016}
Marius Cordts, Mohamed Omran, Sebastian Ramos, Timo Rehfeld, Markus Enzweiler,
  Rodrigo Benenson, Uwe Franke, Stefan Roth, and Bernt Schiele.
\newblock The {Cityscapes} dataset for semantic urban scene understanding.
\newblock In {\em CVPR 2016}.

\bibitem{couprie:2013}
Camille Couprie, Clement Farabet, Laurent Najman, and Yann LeCun.
\newblock Indoor semantic segmentation using depth information.
\newblock In {\em ICLR 2013}.

\bibitem{dai:2017}
Angela Dai, Angel~X. Chang, Manolis Savva, Maciej Halber, Thomas Funkhouser,
  and Matthias Niessner.
\newblock {ScanNet}: {R}ichly-annotated {3D} reconstructions of indoor scenes.
\newblock In {\em CVPR 2017}.

\bibitem{dosovitskiy:2017}
Alexey Dosovitskiy, German Ros, Felipe Codevilla, Antonio Lopez, and Vladlen
  Koltun.
\newblock {CARLA}: {A}n open urban driving simulator.
\newblock In {\em CoRL 2017}.

\bibitem{felzenszwalb:2004}
Pedro~F. Felzenszwalb and Daniel~P. Huttenlocher.
\newblock Efficient graph-based image segmentation.
\newblock {\em IJCV}, 59(2), 2004.

\bibitem{fu:2020}
Huan Fu, Rongfei Jia, Lin Gao, Mingming Gong, Binqiang Zhao, Steve Maybank, and
  Dacheng Tao.
\newblock {3D-FUTURE}: {3D} furniture shape with {TextURE}.
\newblock arXiv 2020.

\bibitem{gaidon:2016}
Adrien Gaidon, Qiao Wang, Yohann Cabon, and Eleonora Vig.
\newblock Virtual worlds as proxy for multi-object tracking analysis.
\newblock In {\em CVPR 2016}.

\bibitem{garcia:2018}
Alberto Garcia-Garcia, Pablo Martinez-Gonzalez, Sergiu Oprea, John~Alejandro
  Castro-Vargas, Sergio Orts-Escolano, Jose Garcia-Rodriguez, and Alvaro
  Jover-Alvarez.
\newblock {The RobotriX}: {A}n extremely photorealistic and very-large-scale
  indoor dataset of sequences with robot trajectories and interactions.
\newblock In {\em IROS 2018}.

\bibitem{geiger:2013}
Andreas Geiger, Philip Lenz, Christoph Stiller, and Raquel Urtasun.
\newblock Vision meets robotics: {T}he {KITTI} dataset.
\newblock {\em IJRR}, 2013.

\bibitem{genova:2017}
Kyle Genova, Manolis Savva, Angel~X. Chang, and Thomas Funkhouser.
\newblock Learning where to look: {D}ata-driven viewpoint set selection for
  {3D} scenes.
\newblock arXiv 2017.

\bibitem{gkioxari:2019}
Georgia Gkioxari, Jitendra Malik, and Justin Johnson.
\newblock Mesh {R-CNN}.
\newblock In {\em ICCV 2019}.

\bibitem{gupta:2013}
Saurabh Gupta, Pablo Arbelaez, and Jitendra Malik.
\newblock Perceptual organization and recognition of indoor scenes from {RGB-D}
  images.
\newblock In {\em CVPR 2013}.

\bibitem{handa:2016b}
Ankur Handa, Viorica Patraucean, Vijay Badrinarayanan, Simon Stent, and Roberto
  Cipolla.
\newblock Understanding real world indoor scenes with synthetic data.
\newblock In {\em CVPR 2016}.

\bibitem{handa:2016a}
Ankur Handa, Viorica Patraucean, Simon Stent, and Roberto Cipolla.
\newblock {SceneNet}: {A}n annotated model generator for indoor scene
  understanding.
\newblock In {\em ICRA 2016}.

\bibitem{handa:2014}
Ankur Handa, Thomas Whelan, John McDonald, and Andrew~J. Davison.
\newblock A benchmark for {RGB-D} visual odometry, {3D} reconstruction and
  {SLAM}.
\newblock In {\em ICRA 2014}.

\bibitem{he:2016}
Kaiming He, Xiangyu Zhang, Shaoqing Ren, and Jian Sun.
\newblock Deep residual learning for image recognition.
\newblock In {\em CVPR 2016}.

\bibitem{hua:2016}
Binh-Son Hua, Quang-Hieu Pham, Duc~Thanh Nguyen, Minh-Khoi Tran, Lap-Fai Yu,
  and Sai-Kit Yeung.
\newblock {SceneNN}: {A} scene meshes dataset with {aNNotations}.
\newblock In {\em 3DV 2016}.

\bibitem{hurl:2019}
Braden Hurl, Krzysztof Czarnecki, and Steven Waslander.
\newblock Precise synthetic image and {LiDAR} ({PreSIL}) dataset for autonomous
  vehicle perception.
\newblock arXiv 2019.

\bibitem{isenburg:2005}
Martin Isenburg and Peter Lindstrom.
\newblock Streaming meshes.
\newblock In {\em Visualization 2005}.

\bibitem{jiang:2018}
Chenfanfu Jiang, Siyuan Qi, Yixin Zhu, Siyuan Huang, Jenny Lin, Lap-Fai Yu,
  Demetri Terzopoulos, and Song-Chun Zhu.
\newblock Configurable {3D} scene synthesis and {2D} image rendering with
  per-pixel ground truth using stochastic grammars.
\newblock {\em IJCV}, 126(9), 2018.

\bibitem{jin:2020}
Lei Jin, Yanyu Xu, Jia Zheng, Junfei Zhang, Rui Tang, Shugong Xu, Jingyi Yu,
  and Shenghua Gao.
\newblock Geometric structure based and regularized depth estimation from 360
  degree indoor imagery.
\newblock In {\em CVPR 2020}.

\bibitem{khan:2019}
Samin Khan, Buu Phan, Rick Salay, and Krzysztof Czarnecki.
\newblock {ProcSy}: {P}rocedural synthetic dataset generation towards influence
  factor studies of semantic segmentation networks.
\newblock In {\em CVPR 2019 Workshops}.

\bibitem{kolve:2017}
Eric Kolve, Roozbeh Mottaghi, Winson Han, Eli VanderBilt, Luca Weihs, Alvaro
  Herrasti, Daniel Gordon, Yuke Zhu, Abhinav Gupta, and Ali Farhadi.
\newblock {AI2-THOR}: {A}n interactive {3D} environment for visual {AI}.
\newblock arXiv 2017.

\bibitem{kovacs:2017}
Balazs Kovacs, Sean Bell, Noah Snavely, and Kavita Bala.
\newblock Shading annotations in the wild.
\newblock In {\em CVPR 2017}.

\bibitem{krahenbuhl:2018}
Philipp Kr\"ahenb\"uhl.
\newblock Free supervision from video games.
\newblock In {\em CVPR 2018}.

\bibitem{li:2016}
Ke Li and Jitendra Malik.
\newblock Amodal instance segmentation.
\newblock In {\em ECCV 2016}.

\bibitem{li:2018a}
Wenbin Li, Sajad Saeedi, John McCormac, Ronald Clark, Dimos Tzoumanikas, Qing
  Ye, Yuzhong Huang, Rui Tang, and Stefan Leutenegger.
\newblock {InteriorNet}: {M}ega-scale multi-sensor photo-realistic indoor
  scenes dataset.
\newblock In {\em BMVC 2018}.

\bibitem{li:2020a}
Zhengqin Li, Mohammad Shafiei, Ravi Ramamoorthi, Kalyan Sunkavalli, and
  Manmohan Chandraker.
\newblock Inverse rendering for complex indoor scenes: {S}hape,
  spatially-varying lighting and {SVBRDF} from a single image.
\newblock In {\em CVPR 2020}.

\bibitem{li:2018b}
Zhengqi Li and Noah Snavely.
\newblock {CGIntrinsics}: {B}etter intrinsic image decomposition through
  physically-based rendering.
\newblock In {\em ECCV 2018}.

\bibitem{li:2020b}
Zhengqin Li, Ting-Wei Yu, Shen Sang, Sarah Wang, Sai Bi, Zexiang Xu, Hong-Xing
  Yu, Kalyan Sunkavalli, Miloš Hašan, Ravi Ramamoorthi, and Manmohan
  Chandraker.
\newblock {OpenRooms}: {A}n end-to-end open framework for photorealistic indoor
  scene datasets.
\newblock arXiv 2020.

\bibitem{lim:2013}
Joseph~J. Lim, Hamed Pirsiavash, and Antonio Torralba.
\newblock Parsing {IKEA} objects: {F}ine pose estimation.
\newblock In {\em ICCV 2013}.

\bibitem{lin:2014}
Tsung-Yi Lin, Michael Maire, Serge Belongie, Lubomir Bourdev, Ross Girshick,
  James Hays, Pietro Perona, Deva Ramanan, C.~Lawrence Zitnick, and Piotr
  Doll{\'a}r.
\newblock {Microsoft COCO}: {C}ommon objects in context.
\newblock In {\em ECCV 2014}.

\bibitem{liu:2019}
Bingyuan Liu, Jiantao Zhang, Xiaoting Zhang, Wei Zhang, Chuanhui Yu, and Yuan
  Zhou.
\newblock Furnishing your room by what you see: {A}n end-to-end furniture set
  retrieval framework with rich annotated benchmark dataset.
\newblock arXiv 2019.

\bibitem{luebke:2002}
David Luebke, Martin Reddy, Jonathan~D. Cohen, Amitabh Varshney, Benjamin
  Watson, and Robert Huebner.
\newblock {\em Level of Detail for {3D} Graphics}.
\newblock Morgan Kaufmann, 2002.

\bibitem{malandain:2002}
Gr{\'e}goire Malandain and Jean-Daniel Boissonnat.
\newblock Computing the diameter of a point set.
\newblock {\em International Journal of Computational Geometry \&
  Applications}, 12(6), 2002.

\bibitem{mccormac:2017}
John McCormac, Ankur Handa, Stefan Leutenegger, and Andrew~J. Davison.
\newblock {SceneNet RGB-D}: {C}an 5{M} synthetic images beat generic {ImageNet}
  pre-training on indoor segmentation?
\newblock In {\em ICCV 2017}.

\bibitem{nguyen:2017}
Duc~Thanh Nguyen, Binh-Son Hua, Lap-Fai Yu, and Sai-Kit Yeung.
\newblock A robust {3D-2D} interactive tool for scene segmentation and
  annotation.
\newblock In {\em Pacific Graphics 2017}.

\bibitem{nikolenko:2019}
Sergey~I. Nikolenko.
\newblock Synthetic data for deep learning.
\newblock arXiv 2019.

\bibitem{richter:2016}
Stephan Richter, Vibhav Vineet, Stefan Roth, and Vladlen Koltun.
\newblock Playing for data: {G}round truth from computer games.
\newblock In {\em ECCV 2016}.

\bibitem{richter:2017}
Stephan~R. Richter, Zeeshan Hayder, and Vladlen Koltun.
\newblock Playing for benchmarks.
\newblock In {\em ICCV 2017}.

\bibitem{ronneberger:2015}
Olaf Ronneberger, Philipp Fischer, and Thomas Brox.
\newblock {U-Net}: {C}onvolutional networks for biomedical image segmentation.
\newblock In {\em MICCAI 2015}.

\bibitem{ros:2016}
German Ros, Laura Sellart, Joanna Materzynska, David Vazquez, and Antonio~M.
  Lopez.
\newblock {The SYNTHIA Dataset}: {A} large collection of synthetic images for
  semantic segmentation of urban scenes.
\newblock In {\em ECCV 2016}.

\bibitem{russakovsky:2015}
Olga Russakovsky, Jia Deng, Hao Su, Jonathan Krause, Sanjeev Satheesh, Sean Ma,
  Zhiheng Huang, Andrej Karpathy, Aditya Khosla, Michael Bernstein,
  Alexander~C. Berg, and Li Fei-Fei.
\newblock {ImageNet} large scale visual recognition challenge.
\newblock {\em IJCV}, 2015.

\bibitem{sadat:2018}
Fatemeh~Sadat Saleh, Mohammad~Sadegh Aliakbarian, Mathieu Salzmann, Lars
  Petersson, and Jose~M. Alvarez.
\newblock Effective use of synthetic data for urban scene semantic
  segmentation.
\newblock In {\em ECCV 2018}.

\bibitem{savva:2017}
Manolis Savva, Angel~X. Chang, Alexey Dosovitskiy, Thomas Funkhouser, and
  Vladlen Koltun.
\newblock {MINOS}: {M}ultimodal indoor simulator for navigation in complex
  environments.
\newblock arXiv 2017.

\bibitem{savva:2019}
Manolis Savva, Abhishek Kadian, Oleksandr Maksymets, Yili Zhao, Erik Wijmans,
  Bhavana Jain, Julian Straub, Jia Liu, Vladlen Koltun, Jitendra Malik, Devi
  Parikh, and Dhruv Batra.
\newblock {Habitat}: {A} platform for embodied {AI} research.
\newblock In {\em ICCV 2019}.

\bibitem{sengupta:2019}
Soumyadip Sengupta, Jinwei Gu, Kihwan Kim, Guilin Liu, David~W. Jacobs, and Jan
  Kautz.
\newblock Neural inverse rendering of an indoor scene from a single image.
\newblock In {\em ICCV 2019}.

\bibitem{settles:2012}
Burr Settles.
\newblock {\em Active Learning}.
\newblock Morgan \& Claypool, 2012.

\bibitem{shah:2017}
Shital Shah, Debadeepta Dey, Chris Lovett, and Ashish Kapoor.
\newblock {AirSim}: {H}igh-fidelity visual and physical simulation for
  autonomous vehicles.
\newblock In {\em Field and Service Robotics 2017}.

\bibitem{shi:2017}
Jian Shi, Yue Dong, Hao Su, and Stella~X. Yu.
\newblock Learning non-{Lambertian} object intrinsics across {ShapeNet}
  categories.
\newblock In {\em CVPR 2017}.

\bibitem{shoeybi:2019}
Mohammad Shoeybi, Mostofa Patwary, Raul Puri, Patrick LeGresley, Jared Casper,
  and Bryan Catanzaro.
\newblock {Megatron-LM}: {T}raining multi-billion parameter language models
  using model parallelism.
\newblock arXiv 2019.

\bibitem{silberman:2012}
Nathan Silberman, Pushmeet Kohli, Derek Hoiem, and Rob Fergus.
\newblock Indoor segmentation and support inference from {RGBD} images.
\newblock In {\em ECCV 2012}.

\bibitem{song:2015}
Shuran Song, Samuel~P. Lichtenberg, and Jianxiong Xiao.
\newblock {SUN RGB-D}: {A} {RGB-D} scene understanding benchmark suite.
\newblock In {\em CVPR 2015}.

\bibitem{song:2017}
Shuran Song, Fisher Yu, Andy Zeng, Angel~X. Chang, Manolis Savva, and Thomas
  Funkhouser.
\newblock Semantic scene completion from a single depth image.
\newblock In {\em CVPR 2017}.

\bibitem{standley:2019}
Trevor Standley, Amir~R. Zamir, Dawn Chen, Leonidas Guibas, Jitendra Malik, and
  Silvio Savarese.
\newblock Which tasks should be learned together in multi-task learning?
\newblock arXiv 2019.

\bibitem{straub:2019}
Julian Straub, Thomas Whelan~Lingni Ma, Yufan Chen, Erik Wijmans, Simon Green,
  Jakob~J. Engel, Raul Mur-Artal, Carl Ren, Shobhit Verma, Anton Clarkson,
  Mingfei Yan, Brian Budge, Yajie Yan, Xiaqing Pan, June Yon, Yuyang Zou,
  Kimberly Leon, Nigel Carter, Jesus Briales, Tyler Gillingham, Elias Mueggler,
  Luis Pesqueira, Manolis Savva, Dhruv Batra, Hauke~M. Strasdat, Renzo~De
  Nardi, Michael Goesele, Steven Lovegrove, and Richard Newcombe.
\newblock {The Replica Dataset}: {A} digital replica of indoor spaces.
\newblock arXiv 2019.

\bibitem{sun:2017}
Xingyuan Sun, Jiajun Wu, Xiuming Zhang, Zhoutong Zhang, Chengkai Zhang, Tianfan
  Xue, Joshua~B. Tenenbaum, and William~T. Freeman.
\newblock {Pix3D}: {D}ataset and methods for single-image {3D} shape modeling.
\newblock In {\em CVPR 2017}.

\bibitem{vasiljevic:2019}
Igor Vasiljevic, Nick Kolkin, Shanyi Zhang, Ruotian Luo, Haochen Wang,
  Falcon~Z. Dai, Andrea~F. Daniele, Mohammadreza Mostajabi, Steven Basart,
  Matthew~R. Walter, and Gregory Shakhnarovich.
\newblock {DIODE}: {A} {D}ense {I}ndoor and {O}utdoor {DE}pth {D}ataset.
\newblock arXiv 2019.

\bibitem{wang:2019}
Qiang Wang, Shizhen Zheng, Qingsong Yan, Fei Deng, Kaiyong Zhao, and Xiaowen
  Chu.
\newblock {IRS}: {A} large synthetic indoor robotics stereo dataset for
  disparity and surface normal estimation.
\newblock arXiv 2019.

\bibitem{wang:2020}
Wenshan Wang, Delong Zhu, Xiangwei Wang, Yaoyu Hu, Yuheng Qiu, Chen Wang, Yafei
  Hu, Ashish Kapoor, and Sebastian Scherer.
\newblock {TartanAir}: {A} dataset to push the limits of visual {SLAM}.
\newblock arXiv 2020.

\bibitem{wrenninge:2018}
Magnus Wrenninge and Jonas Unger.
\newblock {Synscapes}: {A} photorealistic synthetic dataset for street scene
  parsing.
\newblock {\em arXiv 2018}.

\bibitem{wu:2018}
Yi Wu, Yuxin Wu, Georgia Gkioxari, and Yuandong Tian.
\newblock Building generalizable agents with a realistic and rich {3D}
  environment.
\newblock {\em arXiv 2018}.

\bibitem{xia:2018}
Fei Xia, Amir~R. Zamir, Zhi-Yang He, Alexander Sax, Jitendra Malik, and Silvio
  Savarese.
\newblock {Gibson Env}: {R}eal-world perception for embodied agents.
\newblock In {\em CVPR 2018}.

\bibitem{xiang:2016}
Yu Xiang, Wonhui Kim, Wei Chen, Jingwei Ji, Christopher Choy, Hao Su, Roozbeh
  Mottaghi, Leonidas Guibas, and Silvio Savarese.
\newblock A large scale database for {3D} object recognition.
\newblock In {\em ECCV 2016}.

\bibitem{xiang:2014}
Yu Xiang, Roozbeh Mottaghi, and Silvio Savarese.
\newblock Beyond {PASCAL}: {A} benchmark for {3D} object detection in the wild.
\newblock In {\em WACV 2014}.

\bibitem{xiao:2013}
Jianxiong Xiao, Andrew Owens, and Antonio Torralba.
\newblock {SUN3D}: {A} database of big spaces reconstructed using {SfM} and
  object labels.
\newblock In {\em ICCV 2013}.

\bibitem{zhang:2017}
Yinda Zhang, Shuran Song, Ersin Yumer, Manolis Savva, Joon-Young Lee, Hailin
  Jin, and Thomas Funkhouser.
\newblock Physically-based rendering for indoor scene understanding using
  convolutional neural networks.
\newblock In {\em CVPR 2017}.

\bibitem{zheng:2019}
Jia Zheng, Junfei Zhang, Jing Li, Rui Tang, Shenghua Gao, and Zihan Zhou.
\newblock {Structured3D}: {A} large photo-realistic dataset for structured {3D}
  modeling.
\newblock arXiv 2019.

\bibitem{zhou:2019}
Bolei Zhou, Hang Zhao, Xavier Puig, Tete Xiao, Sanja Fidler, Adela Barriuso,
  and Antonio Torralba.
\newblock Semantic understanding of scenes through {ADE20K} dataset.
\newblock {\em IJCV}, 2019.

\end{thebibliography}
}

\end{document}